\pdfoutput=1

\documentclass[11pt]{article}

\usepackage[final]{acl}

\usepackage{times}
\usepackage{latexsym}

\usepackage[T1]{fontenc}

\usepackage[utf8]{inputenc}

\usepackage{microtype}

\usepackage{inconsolata}

\usepackage{graphicx}

\usepackage{MnSymbol}
\usepackage{tabularx}
\usepackage{booktabs}
\usepackage{placeins}

\title{A comparison of data filtering techniques for English-Polish LLM-based machine translation in the biomedical domain}

\author{Jorge del Pozo Lérida \\
  IT University of Copenhagen \\
  \texttt{jord@itu.dk} \\\And
  Kamil Kojs \\
  IT University of Copenhagen \\
  \texttt{kako@itu.dk} \\\AND
  János Máté \\
  IT University of Copenhagen \\
  \texttt{janma@itu.dk}
  \And
  Mikołaj Antoni Barański \\
  IT University of Copenhagen \\
  \texttt{mikba@itu.dk} \\
  \And
  Christian Hardmeier \\
  IT University of Copenhagen \\
  \texttt{chrha@itu.dk} \\
}

\begin{document}
\maketitle
\begin{abstract}
Large Language Models (LLMs) have become state-of-the-art in Machine Translation (MT), often trained on massive bilingual parallel corpora scraped from the web, that contain low-quality entries and redundant information, leading to significant computational challenges. Various data filtering methods exist to reduce dataset sizes, but their effectiveness largely varies based on specific language pairs and domains. This paper evaluates the impact of commonly used data filtering techniques—LASER, MUSE, and LaBSE—on English-Polish translation within the biomedical domain. By filtering the UFAL Medical Corpus, we created varying dataset sizes to fine-tune the mBART50 model, which was then evaluated using the SacreBLEU metric on the Khresmoi dataset, having the quality of translations assessed by bilingual speakers. Our results show that both LASER and MUSE can significantly reduce dataset sizes while maintaining or even enhancing performance. We recommend the use of LASER, as it consistently outperforms the other methods and provides the most fluent and natural-sounding translations. 
\end{abstract}

\section{Introduction}
Recent advancements in LLMs have resulted in a notable increase in the size of model architectures used for MT, with publicly accessible models such as mBART-50 reaching parameter sizes up to 600 million \cite{mBartParameterSize}. This escalation in parameter size has consequently resulted in an increased demand for computational resources, necessitating the use of the most advanced GPUs for training these models. A common approach to training LLMs is to utilize the entirety of the data present in the prepared training dataset. However, it is often the case that these datasets are not entirely free of low-quality entries. Since training datasets frequently comprise over one million samples, often scraped from the web, it is impractical to manually inspect and purify the entire dataset of such substandard samples. Consequently, it is typical that only a minimal fraction of the dataset is of high quality, with the majority being of poor quality.

Past research has shown that including low-quality data in the training process minimally contributes to the overall quality and performance of LLMs \cite{koehn-etal-2018-findings}, underscoring the imperative to refine training data sets to contain only high-quality entries. In this study, we hypothesize that the application of filtering techniques can significantly reduce the size of the training dataset in MT with LLMs, when fine-tuning for a specific domain and language pair, without compromising or potentially improving performance.

Specifically, we investigate domain adaptation for biomedical translation from English to Polish, performing a systematic comparison of common filtering methods to identify the best candidate for this particular setup. We achieve this by filtering a large in-domain biomedical corpus into smaller datasets of different sizes using different methods. We then fine-tune the mBART50 model on these filtered subsets and evaluate it on an independent dataset, comparing performance to that of using a whole corpus or randomly sampled equally sized subsets. Our main contribution is to provide the first systematic comparison of data filtering methods for English-Polish biomedical MT and to provide specific recommendations. For the sake of reproducibility, all code is made available. 
\footnote{\href{https://github.com/jorgedelpozolerida/Biomed-NMT-EngPol}{https://github.com/jorgedelpozolerida/Biomed-NMT-EngPol}}

\section{Related Work}
\label{sec:rel_works}
\begin{table*}[ht]
\centering
\small 
\begin{tabular}{@{}lcccccc@{}} 
\toprule
\textbf{Model} & \textbf{Filter} & \textbf{Avg. BLEU} & \textbf{Min} & \textbf{Max} & \textbf{$\Delta$ All} & \textbf{$\Delta$ 60/20} \\
\midrule
Base-none & - & 14.936 & - & - & -2.466 & - \\
Base-all & - & 17.402 & - & - & - & - \\
Base-60\% & - & 17.234 & 17.174 & 17.296 & -0.168 & - \\
Filtered-60\% & LASER & \textbf{17.411} & - & - & \textbf{0.009} & \textbf{0.177} \\
Filtered-60\% & MUSE & 17.239 & - & - & -0.163 & 0.005 \\
Filtered-60\% & LaBSE &  17.151 & - & - & -0.251 & -0.083 \\
Base-20\% & - & 16.801 & 16.516 & 17.041 & -0.601 & - \\
Filtered-20\% & LASER & \textbf{17.114} & - & - & \textbf{-0.288} & \textbf{0.313} \\
Filtered-20\% & MUSE & 17.071 & - & - & -0.331 & 0.270 \\
Filtered-20\% & LaBSE & 16.376 & - & - & -1.026 & -0.425 \\
\bottomrule
\end{tabular}
\caption{\label{tab:experiment-results} Evaluation results on the Khresmoi test dataset using SacreBLEU. For the Base-60\% and Base-20\% models average scores between random seeds are reported, with standard deviations of 0.05 and 0.13 respectively.}
\end{table*}




LLM-based MT models depend on large quantities of high-quality data for domain adaptation \cite{koehn-etal-2018-findings}. To enhance the quality of web-scraped corpora, various automated filtering methods have been explored within the MT field, including outlier detection \cite{taghipour2011parallel}, discriminator models \cite{xu2017zipporah}, graph-based unsupervised models \cite{cui2013bilingual}, and LLM-based classifiers or scorers \cite{accarcciccek2020filtering}. With the advent of LLMs, language-agnostic encoders such as LASER, LaBSE, and MUSE have enabled the direct scoring of bilingual sentence similarity for data set filtering, proving competitive with more complex classifier-based models \cite{chaudhary2019low}. 

The research by \citet{bane2021selecting} evaluated the effectiveness of filtering in English-Japanese and English-German sentence pairs using models like Marian-scorer, LASER, MUSE and XLM-R. The findings showed limited reduction of the dataset (54\%-73\%) with BLEU scores comparable to random selection, although Marian-based filtering consistently outperformed random downsizing, and MUSE showed variable performance by language. The latter indicates that filtering results might not be universally applicable across different pairs or topics of languages, which motivated our study design to specifically investigate the Polish-English translation.

Further exploration by \citet{bane2022comparison} assessed the strengths and weaknesses of specific filtering methods. They developed a dataset with ten types of noise or errors to test these methods. Results indicated that a custom-trained Marian-Scorer had the best cleaning performance, while embedding-based methods like XLM-R, MUSE, and LASER, although less effective, still performed adequately and were particularly effective at identifying issues like number mismatches and spelling errors without requiring the computationally costly calibration needed for the Marian-Scorer.

\section{Data}
\label{sec:data}
The selected model was fine-tuned on the Polish-English sentence pairs from the UFAL Medical Corpus \footnote{\href{https://ufal.mff.cuni.cz/ufal_medical_corpus}{https://ufal.mff.cuni.cz/ufal\_medical\_corpus}}, consisting of 1,116,773 pairs sourced from documents of the European Centre for Disease Prevention and Control, the European Medicines Agency, and Open subtitles. Extensive preprocessing included removing duplicates, untranslated sentences, those under 15 or over 200 characters, and sentences containing characters from non-target languages, resulting in a refined dataset of 700,00 sentence pairs. For testing, we used the Khresmoi dataset \citep{khresmoi_summary_translation_test_data_2.0}, which comprises 1,500 high-quality Polish-English medical sentence pairs.

\section{Methodology}
We employed three widely used multilingual embedding models — LASER, MUSE, LaBSE — to help us filter the medical-domain corpus. Each embedding method was used to generate a sentence representation of each sentence in a pair, either by averaging all token embeddings in the sentence or by taking the sentence representation from the method if already provided. We then utilized cosine similarity to score sentence pairs from the in-domain training data, retaining 20\% (approximately 150k pairs) and 60\% (approximately 420k pairs) of the highest-scoring sentences from each method to create eight different filtered training datasets for all combinations of embedding methods and sizes. Each subset was used to separately fine-tune a pre-trained model, specifically mBART50, which was selected due to its public availability and for its good performance in our language pair, as well as its reasonable size for conducting experiments. Evaluation was conducted on an independent dataset. Additionally, two authors of this paper, KK and MB, who are native Polish speakers and proficient in English, qualitatively assessed the translations to determine which method produced more natural-sounding results.



\subsection{Filtering methods}

\subsubsection{LaBSE}
LaBSE (Language-agnostic BERT Sentence Embeddings) is a method developed by Google for generating BERT-based cross-lingual sentence embeddings in over 109 languages \cite{feng2022languageagnostic}. This model, available on Hugging Face \footnote{\href{https://huggingface.co/sentence-transformers/LaBSE}{https://huggingface.co/sentence-transformers/LaBSE}}, addresses the limitations of the original BERT’s multilingual embeddings by employing a dual-encoder framework. It uses a pre-trained BERT to produce embeddings separately for each sentence in the translated pair, with the training loss calculated as the difference between these embeddings, facilitating the development of a unified cross-lingual embedding space. LaBSE was chosen for our analysis due to its demonstrated potential in new applications and its effectiveness in cross-lingual settings, and because it has not been studied in previous data filtering literature.


\subsubsection{LASER}

LASER (Language-Agnostic SEntence Representations) \cite{LASER}, developed by Facebook AI Research, utilizes a BiLSTM (Bidirectional Long Short-Term Memory) architecture to create language-agnostic sentence embeddings \footnote{\href{https://github.com/facebookresearch/LASER}{https://github.com/facebookresearch/LASER}}. This model is trained on a large multilingual dataset of parallel corpora, enabling it to generate consistent embeddings for semantically equivalent sentences across more than 90 languages. LASER employs a Byte Pair Encoding (BPE) tokenizer to process various languages by segmenting words into shared subword units, enhancing its language-generalization capability. It encodes input sentences into a fixed-size vector. Due to its robust performance in data filtering across different domains and languages, LASER was selected for our analysis \cite{chaudhary2019low}.


\subsubsection{MUSE}
MUSE (Multilingual Unsupervised and Supervised Embeddings) is a model developed by Meta to foster the creation and evaluation of cross-lingual word embeddings \cite{conneau2017word}. This model uses an unsupervised approach for aligning monolingual word embeddings, which includes adversarial training to establish a linear mapping between source and target embedding spaces, synthesizing a dictionary from the mapped space, and refining the alignment with the Procrustes solution, allowing for cross-lingual alignment without annotated data or parallel corpora. Though the original model is no longer available, we utilized its pre-compiled embeddings dictionary \footnote{\href{https://ai.meta.com/tools/muse/}{https://ai.meta.com/tools/muse/}}. MUSE was selected for our research due to its demonstrated effectiveness across various language pairs and domains in data filtering tasks \cite{bane2021selecting}.

\begin{table*}
\tiny
\begin{tabularx}{\textwidth}{lX} 
\toprule
\textbf{Type} & \textbf{Sentence} \\
\midrule
\textbf{English} & Meningococcal Disease is a serious bacterial infection that can cause swelling of the brain and spinal cord, and infection of the blood and other organs. \\
\textbf{Ground Truth} & Infekcja meningokokowa jest poważną chorobą bakteryjną, która może spowodować obrzęk mózgu i rdzenia, infekcję krwi i innych narządów. \\
\midrule
\textbf{LaBSE-60} & Meningooka jest ciężkim zakażeniem bakteryjnym, które może powodować obrzęk mózgu i rdzenia kręgowego oraz zakażenie krwi i innych narządów. \\
\textbf{LASER-60} & Choroba meningokokowa jest ciężkim zakażeniem bakteryjnym, które może powodować obrzęk mózgu i rdzenia kręgowego oraz zakażenie krwi i innych narządów. \\
\textbf{MUSE-60} & Choroba meningokokowa jest ciężkim zakażeniem bakteryjnym, które może powodować obrzęk mózgu i rdzenia kręgowego oraz zakażenie krwi i innych narządów. \\
\textbf{Base-60} & Choroba meningokokowa jest ciężkim zakażeniem bakteryjnym, które może powodować obrzęk mózgu i rdzenia kręgowego oraz zakażenie krwi i innych narządów. \\
\midrule
\textbf{LaBSE-20} & Meningokoczka jest poważnym zakażeniem bakteryjnym, które może powodować obrzęk mózgu i rdzenia kręgowego oraz zakażenie krwi i innych narządów. \\
\textbf{LASER-20} & Choroba meningokokowa jest poważnym zakażeniem bakteryjnym, które może powodować obrzęk mózgu i rdzenia kręgowego oraz zakażenie krwi i innych narządów. \\
\textbf{MUSE-20} & Choroba gruczołu krokowego jest poważnym zakażeniem bakteryjnym, które może powodować obrzęk mózgu i rdzenia kręgowego oraz zakażenie krwi i innych narządów. \\
\textbf{Base-20} & Choroba meningokokowa jest ciężkim zakażeniem bakteryjnym, które może powodować obrzęk mózgu i rdzenia kręgowego oraz zakażenie krwi i innych narządów. \\
\midrule
\textbf{Base-all} & Choroba meningokokowa jest ciężkim zakażeniem bakteryjnym, które może powodować obrzęk mózgu i rdzenia kręgowego oraz zakażenie krwi i innych narządów. \\
\textbf{Base-none} & Chorób gruczołu krokowego jest poważnym zakażeniem bakteryjnym, które może powodować obrzęk mózgu i rdzenia kręgowego, i zakażenie krwi i innych organów. \\
\bottomrule
\end{tabularx}
\caption{\label{tab:trans1} Example of evaluation translations for different models. Only one seed is shown for Base-60 and Base-20.}
\end{table*}

\subsection{Experimental setup}
\label{sec:exp_setup}

We compared the performance of 10 models, with 9 of them fine-tuned on variously sized subsets of UFAL Medical Corpus and one remaining being the untouched pre-trained baseline (\textit{Base-none}). The latter was fine-tuned on the full unfiltered dataset to obtain a fine-tuned baseline (\textit{Base-all}). We employed a stratified split of 80\% training and 20\% validation for all experiments to maintain consistent proportions of the three different data sources. To be able to observe the effect of filtering against randomness, baseline models were trained on randomly selected subsets of 20\% and 60\% (\textit{Base-20\%} and \textit{Base-60\%}), each trained three times with different seeds to average out random variance in performance evaluation.


We utilized the publicly available mBART50 model \citep{tang2020multilingual}\footnote{\href{https://huggingface.co/facebook/mbart-large-50}{https://huggingface.co/facebook/mbart-large-50}}, developed by Facebook AI and available for use in Polish and English, with text tokenization performed by the MBart50Tokenizer. Evaluation was conducted on an independent test dataset using the BLEU metric implemented via SacreBLEU \cite{sacreBleu}, ensuring unbiased assessment. Training time was also reported to highlight the efficiency gains from fine-tuning on smaller data subsets.

The calculations were performed on the LUMI supercomputer \footnote{\href{https://www.lumi-supercomputer.eu/about-lumi}{https://www.lumi-supercomputer.eu/about-lumi}}, using its \textit{ standard g} partition with AMD MI250x GPUs, totaling 646 GPU hours. Training involved using Trainer function from transformer library, with 16-bit precision for weights, batch sizes of 15 for training and 20 for evaluation, a linear learning rate scheduler for the initial 100 steps, and an AdamW optimizer. All models underwent exactly three training epochs. More training details can be seen in Table \ref{tab:model_types}.

\section{Results}
Unsurprisingly, when looking at evaluation results in Table~\ref{tab:experiment-results} we observe that fine-tuning on in-domain data improves performance, as \textit{Base-none} shows a worse performance than any of the fine-tuned models, whereas \textit{Base-all} shows a BLEU of 17.402, a 2.466 increase from \textit{Base-none}. When training on smaller random subsets of the data, \textit{Base-20\%} and \textit{Base-60\%} show less performance increase than when using the whole corpus, as was
expected. 

The benefit of non-random filtering is especially visible in \textit{LASER-60\%}, where performance is higher than \textit{Base-60\%} (an increase of 0.177) and even than \textit{Base-all} (an increase of 0.009), meaning that removing 40\% of the "worst quality data" yielded marginally increased performance. The case of \textit{MUSE-60\%} is also positive since it only meant a decrease of 0.163 compared to \textit{Base-all}, and performance was higher than \textit{Base-60\%} by 0.005. The case of \textit{LaBSE-60\%} is different, since it decreased the performance by 0.083 compared to \textit{Base-60\%}, indicating its use was not beneficial. These results are reflected in the manual verification of test translations, where \textit{LASER-60\%} together with \textit{Base-all} produces the most accurate and well-sounding translations (see example in Table \ref{tab:trans1}).

When comparing the smaller sizes of subsets of 20\% of the data, none of the filtering methods helped obtain a model that was better than \textit{Base-all}, but we observe that LASER and MUSE outperform \textit{Base-20\%}. Our human qualitative assessment of translations showed that \textit{MUSE-20\%} struggles with medical terminology, while \textit{LASER-20\%} produces consistently high-quality and natural-sounding text. Here again, the use of LaBSE is not beneficial.

\section{Discussion}

Altogether, our evaluation of data filtering methods on English to Polish translations in the biomedical domain reveals a performance hierarchy: \textbf{LASER\textsubscript{n} > MUSE\textsubscript{n} > Baseline\textsubscript{n} > LaBSE\textsubscript{n}}, with \textbf{n} indicating the subset size. LASER proved to be the most effective, enhancing performance even more than the full corpus when using only 60\% of the data, reducing computing time by nearly half. Furthermore, when using 20\% of the data,  LASER and MUSE achieved relatively lower validation BLEU scores (-0.288 and -0.331 respectively) compared to the baseline model (\textit{Base-all}), but significantly better than the unfiltered baseline model (\textit{Base-none}), with scores of 2.178 and 2.135 respectively. However, \textit{MUSE-20\%}'s translations appear less accurate in specialized medical terminology based on our qualitative inspection, e.g. by losing the real meaning as is seen in Table \ref{tab:trans1}. This was partly expected, as short sentences full of medical terminology were given a low score by the MUSE filtering.    

Surprisingly, LaBSE, expected to perform comparably to LASER, did not meet expectations despite high score correlations (\textit{r} = 0.81) between the two methods. Differences in scoring specific sentences might explain LASER's superior performance. In summary, our findings validate the efficiency of LASER in reducing dataset size without compromising, and sometimes enhancing, model performance, thus affirmatively answering our research questions \textit{R1} and \textit{R2}. MUSE, while effective, was less consistent in translation quality. LaBSE, despite its expected potential, fell short in this specific setting. 

Overall, we recommend LASER as the most effective data filtering method for LLM-based machine translation from English to Polish in the biomedical domain.

\begin{table*}[ht]
\centering
\begin{tabular}{ccccccc}
\hline
\textbf{Dataset} & \textbf{Size} & \textbf{Training Time} & \textbf{Baseline} & \textbf{LaBSE} & \textbf{MUSE} & \textbf{LASER} \\
\hline
Base-none & n/a & n/a & $\checkmark$ & & & \\
Base-all & 700k & 17H:20M & $\checkmark$ & & & \\
Base-60\% & 420k & 10H:30M & 3 seeds & & & \\
Filtered-60\% & 426k & 11H:00M & & $\checkmark$ & $\checkmark$ & $\checkmark$ \\
Base-20\% & 150k & 03H:05M & 3 seeds & & & \\
Filtered-20\% & 158k & 03H:20M & & $\checkmark$ & $\checkmark$ & $\checkmark$ \\
\hline
\end{tabular}
\caption{\label{tab:model_types}Model specifications and average training times. \textit{Base-none} is the raw pre-trained model without any fine-tuning. \textit{Base-all}, \textit{Base-20\%}, and \textit{Base-60\%} are models fine-tuned on all, randomly selected 20\%, and randomly selected 60\% of the training data respectively. Equivalently, the filtered models were fine-tuned on subsets of the data selected as the highest-scored sentence pairs for each filtering method. The reported training times are based on 3 epochs of training on the LUMI supercomputer. Their values are indicative based on representative training runs.}
\end{table*}

\section*{Limitations}

A primary limitation of our study is that all models were trained for only 3 epochs, with results reported for the final model state. This approach may not fully capture the potential of the models if they were subjected to more or fewer training epochs. Future work could try instead to vary the number of epochs and perhaps use a different stopping criterion to explore how it impacts model performance and efficiency, particularly assessing whether fewer epochs could suffice in achieving optimal results with filtered data, thus optimizing training resources.

Another limitation of this study is that our filtering process relied solely on cosine similarity to evaluate semantic similarity between sentence pairs. Other similarity scores could be investigated for this task. Moreover, no filtering method that assessed the domain specificity of the sentences was used in the study, which led to the inclusion of sentences that might not be entirely pertinent to the medical domain, e.g. because only contained some medical proper nouns. It may also be beneficial to complement BLEU score with other metrics such as COMET and to use statistical methods to enhance the significance of the results, such as bootstrapping or using multiple train/dev splits on the filtered subsets used for fine-tuning.

Additionally, our analysis did not include any quantitative human evaluation of model predictions by a translation expert. This would involve selecting top predictions from models fine-tuned on both filtered and unfiltered datasets and having an expert assess them without knowledge of their origin to provide unbiased quality evaluations. Furthermore, expanding the sample size of tested models and implementing significance testing would be needed to fully bolster the robustness and generalizability of our results, offering a more detailed understanding of the models' performance across various settings.
\bibliographystyle{acl_natbib}

\end{document}